\documentclass[11pt, a4paper, logo, copyright]{daios}

\usepackage[sort&compress]{natbib}
\usepackage{amsmath, amssymb}
\usepackage{enumitem}
\setlist[itemize]{topsep=0pt}

\title{Notes on the Reward Representation of Posterior Updates}
\correspondingauthor{pedro@daios.ai}
\paperurl{Technical Report}
\reportnumber{001}
\renewcommand{\today}{February 2026}

\author[*]{Pedro A. Ortega}
\affil[*]{Daios Technologies}

\begin{abstract}
Many ideas in modern control and reinforcement learning treat decision-making as inference: start from a baseline distribution and update it when a “signal” arrives. We ask when this can be made literal rather than metaphorical. We study the special case where a KL-regularized soft update is exactly a Bayesian posterior inside a single fixed probabilistic model, so the update variable is a genuine channel through which information is transmitted. In this regime, behavioral change is driven only by evidence carried by that channel: the update must be explainable as an evidence reweighting of the baseline. This yields a sharp identification result: posterior updates determine the relative, context-dependent incentive signal that shifts behavior, but they do not uniquely determine absolute rewards, which remain ambiguous up to context-specific baselines. Requiring one reusable continuation value across different update directions adds a further coherence constraint linking the reward descriptions associated with different conditioning orders.
\end{abstract}

\begin{document}
\maketitle

\section{Introduction and contributions}

KL-regularized (or bounded-rational) optimization is a standard device in maximum-entropy control, variational inference, and regularized reinforcement learning \citep{ortegabraun2013thermo,ortega2015itbr,wainwrightjordan2008gm,geist2019regularizedmdp}.
In this note we focus on a sharp identification question: what changes when we \emph{restrict} to the subclass of KL-regularized objectives whose optimizer is \emph{exactly} a Bayes posterior conditional under a \emph{single} ambient joint distribution?

This is not a generic property, but a structural constraint on admissible reward/value families: in the identified class, the dependent interaction must coincide with a pointwise mutual information (PMI). More concretely, we impose two explicit assumptions beyond KL regularization. First, \emph{posterior
identification}: the optimizer equals the posterior conditional under the same
joint distribution. Second, \emph{order-independence}: across update directions, a single value function is used consistently, so continuation depends only on the information event (equivalently, the set of conditioned variables). These assumptions force three outcomes:
\begin{itemize}
\item \textbf{Representation:} posterior identification pins the update log-ratio
pointwise; equivalently, it yields a posterior log-ratio representation (conditional PMI) for the identified interaction.
\item \textbf{Identifiability limit:} posteriors identify only a context-baseline-invariant
reward--value interaction (a gauge class), not rewards uniquely without a baseline convention.
\item \textbf{Coherence constraint:} order-independence imposes a commutativity (integrability)
constraint across update directions, coupling the reward parametrizations for different conditioning orders.
\end{itemize}
In essence, posterior identification yields a pointwise representation structure: the only compatible reward--value interaction is the posterior log-ratio, i.e.\ conditional PMI, with a residual context-baseline gauge.

\paragraph{Why this matters.}
Many ``control as inference'' and generalized-Bayes constructions produce \emph{posterior-like} soft updates \citep{bissiri2016generalupdate,wainwrightjordan2008gm,kappen2009gmi_arxiv,levine2018rci}; here we isolate the boundary case where the update is \emph{literally} a Bayes conditional under one ambient $P$. In that case, observed conditionals identify only the interaction that drives behavioral change (a PMI signal), and they cannot by themselves identify absolute reward levels without an external baseline convention \citep{ng1999shaping,kim2021rewardid,cao2021identifiabilityirl}. The commutativity condition then characterizes when a single event-based continuation value can be reused coherently across conditioning orders \citep{ziebart2010mce,ziebart2013mce}.

\section{Main results and roadmap}

We study KL-regularized updates and ask what additional structure is
forced when the optimizer is \emph{identified} with a Bayes posterior conditional
under a single ambient joint distribution. We state the main identities up front
and then derive them in sequence.

\paragraph{Collected identities (under stated assumptions).}
When the corresponding assumptions are in force (defined in Section~\ref{sec:setup}),
the reward/value family is constrained pointwise as follows:
\begin{itemize}
\item \textbf{Posterior PMI representation (equation \ref{eq:pmi-shape}):}
posterior identification forces the reward--value interaction to equal the posterior log-ratio, i.e.\ conditional PMI, pointwise,
\[
\alpha\bigl[r_z(x\mid y)+V(x,y,z)-V(y,z)\bigr]=i(x;z\mid y).
\]
\item \textbf{Gauge freedom (equation \ref{eq:gauge}):} only the interaction is identified; reward-value splits are ambiguous up to context-only baselines.
For any $c(y,z)$, shifting $r_z(x\mid y)\mapsto r_z(x\mid y)+c(y,z)$ and
$V(y,z)\mapsto V(y,z)+c(y,z)$ leaves $P(x\mid y,z)$ unchanged.
\item \textbf{Commutativity / integrability (equation \ref{eq:commute}):}
requiring a single value function across update directions couples reward parametrizations across conditioning orders,
\[
r_z(x\mid y)-V(y,z)=r_x(z\mid y)-V(x,y).
\]
\end{itemize}
The remainder of the note derives these identities and interprets their
implications for inference--optimization unification, the interpretation of agent design, and the limits of reward identifiability from observed conditionals.

\section{Setup}
\label{sec:setup}

\paragraph{Ambient model.}
Fix a joint distribution $P(x,y,z)$ over discrete random variables (or groups of random variables) $X,Y,Z$. We chose three because they're necessary to derive the algebraic results. All conditional/marginal notation is induced by this same $P$. We work pointwise on contexts where the required conditionals are defined.

\paragraph{Value function.}
Values are functions of information states (events): there exists an event-function $\mathcal V$ such that $V(x,y)=\mathcal V(\{X=x,Y=y\})$, $V(x,y,z)=\mathcal V(\{X=x,Y=y,Z=z\})$, etc. Consequently these are permutation-invariant (e.g.\ $V(x,y)=V(y,x)$ and $V(x,y,z)=V(z,y,x)$) by
equality of the underlying events. As is standard in control, the optimization problem treats $V(x,y,z)$ (i.e. with all random variables resolved) as exogenously specified terminal values (a boundary condition) and the internal values such as $V(y,z)$ (i.e. with leftover uncertainty) as derived.

\paragraph{Conditional PMI.}
For any $x,y,z$ with well-defined conditionals under $P$, define the conditional
pointwise mutual information (PMI)
\begin{equation}
\label{eq:pmi}
    i(x;z\mid y)
    := \log \frac{P(x,z\mid y)}{P(x\mid y)\,P(z\mid y)}
    = \log\frac{P(x\mid y,z)}{P(x\mid y)}.
\end{equation}
This is the pointwise analogue of conditional mutual information and inherits standard identities (e.g.\ symmetry in $x,z$ given $y$) \citep{shannon1948communication,coverthomas2006eit}.

\paragraph{Assumptions used in the sequel.}
\begin{enumerate}[leftmargin=*]
\item \textbf{Posterior identification (A1):} when we optimize a KL-regularized
objective whose reference is induced by the ambient joint $P$, we additionally
identify the optimizer with a genuine posterior conditional under the same $P$.
\item \textbf{Coherence across update directions (A2):} when we impose posterior
identification for multiple update directions (e.g.\ updating $X$ given $(Y,Z)$
and updating $Z$ given $(Y,X)$), we require \emph{event-based coherence} encoded in the permutation-invariance of the value function: values depend only on what is known, not on the update direction or the order in which information is acquired.
\end{enumerate}

\section{KL-regularized conditional update}

Fix $\alpha>0$. For any real-valued functions $r_z(x\mid y)$ and terminal values $V(x,y,z)$, define the KL-regularized objective, for each context $(y,z)$,
\begin{equation}
\label{eq:Jw}
    J_z(\tilde{P}(x\mid y))
    := \sum_x \tilde{P}(x\mid y)\Bigl[
        r_z(x\mid y)
        -\frac{1}{\alpha}\log\frac{\tilde{P}(x\mid y)}{P(x\mid y)}
        + V(x,y,z)
    \Bigr],
\end{equation}
where $\tilde{P}(x\mid y)$ ranges over distributions absolutely continuous with
respect to $P(x\mid y)$. Here, $r_z(x\mid y)$ corresponds to the reward of~$x$ given~$y$ obtained through ``applying~$z$''. For now, $z$ can be thought of as an instrument for performing the optimization calculation. Objectives of this free-energy / KL-penalized form are standard in information-theoretic bounded rationality and variational KL duality \citep{ortegabraun2013thermo,ortega2015itbr,csiszar1975idivergence,donskervaradhan1975asymptoticII,wainwrightjordan2008gm}.

The maximizer of \eqref{eq:Jw} exists, is unique, and has the exponential-tilt
form
\begin{equation}
\label{eq:tilt}
    \tilde{P}^\star(x\mid y)
    =
    \frac{
        P(x\mid y)\exp\Bigl\{\alpha\bigl[r_z(x\mid y)+V(x,y,z)\bigr]\Bigr\}
    }{
        \sum_{\tilde{x}} P(\tilde{x}\mid y)\exp\Bigl\{\alpha\bigl[
            r_z(\tilde{x}\mid y)+V(\tilde{x},y,z)
        \bigr]\Bigr\}
    }.
\end{equation}
Moreover, the optimal value equals the log-normalizer
\begin{equation}
\label{eq:Vvw_def}
    V(y,z)
    =
    \max_{\tilde{P}(x\mid y)} J_z(\tilde{P}(x\mid y))
    =
    \frac{1}{\alpha}\log \sum_x P(x\mid y)\exp\Bigl\{
        \alpha\bigl[r_z(x\mid y)+V(x,y,z)\bigr]
    \Bigr\}.
\end{equation}
This exponential-tilt and log-partition structure under KL regularization is classical (e.g.\ relative-entropy control, linearly-solvable control, and variational free energy) \citep{todorov2006lmdp,todorov2008duality,csiszar1975idivergence,donskervaradhan1975asymptoticII,hartmann2017freeenergy}.
We use this standard fact as an algebraic starting point; the representation and identifiability results begin once we impose posterior identification.

\section{Posterior identification pins PMI-shape}

\paragraph{Assumption (Posterior identification).}
We identify the optimizer \eqref{eq:tilt} with a posterior conditional under the
same ambient joint $P$:
\begin{equation}
\label{eq:post-id}
    P(x\mid y,z) := \tilde{P}^\star(x\mid y).
\end{equation}
This is an additional modeling assumption (not a generic property of
KL-regularization): in general, KL-regularized objectives induce ``posterior-like'' updates without coinciding with a Bayes conditional of a single fixed joint distribution \citep{bissiri2016generalupdate,wainwrightjordan2008gm}. The optimization instrument $Z=z$ thus becomes a genuine random variable we condition on, assumed to exist in an encompassing ambient distribution. It is most naturally motivated by control-as-inference views where optimal policies are represented as posteriors (or variational posteriors) in an auxiliary model \citep{kappen2009gmi_arxiv,levine2018rci}, here strengthened to exact identification under one ambient joint $P$.

\paragraph{Identified interaction.}
Under \eqref{eq:post-id}, the update \eqref{eq:tilt} implies the pointwise
log-ratio identity
\begin{equation}
\label{eq:logratio}
    \log\frac{P(x\mid y,z)}{P(x\mid y)}
    =
    \alpha\Bigl[r_z(x\mid y)+V(x,y,z)-V(y,z)\Bigr].
\end{equation}
Using \eqref{eq:pmi}, this is equivalently the PMI-shape identity
\begin{equation}
\label{eq:pmi-shape}
    \alpha\Bigl[r_z(x\mid y)+V(x,y,z)-V(y,z)\Bigr]
    = i(x;z\mid y),
\end{equation}
valid pointwise on positive-mass contexts.

\paragraph{Identifiability and gauge structure.}
Posterior identification determines only the \emph{relative} action preferences induced by the soft update; it does \emph{not} by itself fix a unique split into ``reward'' and ``value.'' Operationally, when comparing two actions $x$ and $x'$ in a fixed context $(y,z)$, the update depends on the difference
\[
\bigl(r_z(x\mid y)+V(x,y,z)\bigr)-\bigl(r_z(x'\mid y)+V(x',y,z)\bigr),
\]
so any $x$-independent shift in the combined signal has no effect on the resulting conditional.
Equivalently, for any function $c(y,z)$, the transformation
\begin{equation}
\label{eq:gauge}
    r_z(x\mid y)\mapsto r_z(x\mid y)+c(y,z),
    \qquad
    V(y,z)\mapsto V(y,z)+c(y,z)
\end{equation}
leaves the optimizer (and hence $P(x\mid y,z)$) unchanged. Thus rewards are identifiable only up to a context-only baseline: posteriors fix the within-context differences of $r_z(x\mid y)+V(x,y,z)$ across $x$, but not an absolute level without an additional calibration convention.
This is a context-dependent analogue of policy-invariant reward transformations and modern reward identifiability obstructions in (entropy-regularized) inverse RL \citep{ng1999shaping,kim2021rewardid,cao2021identifiabilityirl}.

\section{PMI-restated objective}

Equations \eqref{eq:logratio}--\eqref{eq:pmi-shape} suggest using the PMIs as reward proxies that identify an entire equivalence class of reward-value splits. To see this, rearrange \eqref{eq:pmi-shape} to obtain a pointwise decomposition
\[
    r_z(x\mid y)+V(x,y,z)=V(y,z)+\frac{1}{\alpha}i(x;z\mid y).
\]
Substituting into \eqref{eq:Jw} gives, for each context $(y,z)$,
\begin{equation}
\label{eq:pmi-obj}
    J_z(\tilde{P}(x\mid y))
    = V(y,z)
        + \sum_x \tilde{P}(x\mid y)\Bigl[
            \frac{1}{\alpha}i(x;z\mid y)
            -\frac{1}{\alpha}\log\frac{\tilde{P}(x\mid y)}{P(x\mid y)}
        \Bigr].
\end{equation}
Thus, up to the context-only term $V(y,z)$ (which does not affect the optimizer),
the numerically relevant ``reward-like'' signal is
$\frac{1}{\alpha}i(x;z\mid y)$. That is, by fixing the values $i(x;z\mid y)$, we obtain a construction method for building the posterior $P(x|y,z)$ from $P(x|y)$.

\paragraph{Constraints on admissible PMI signals.}
Not every family of functions $i(\cdot;z\mid y)$ is admissible. Because
$i(x;z\mid y)$ must be consistent with a posterior log-ratio induced by a \emph{single} ambient joint $P$, it must satisfy the normalization and compatibility constraints implied by Bayes conditioning. In particular, for each $(y,z)$,
\[
    \log \sum_x P(x\mid y)\exp\{i(x;z\mid y)\}=0,
\]
so that $P(x\mid y,z)=P(x\mid y)\exp\{i(x;z\mid y)\}$ is a valid conditional.
Moreover, across different update directions, order-independence (A2) further
couples these PMI-shaped interactions through the commutativity constraint
\eqref{eq:commute}.

\section{Order-independence yields a commutativity constraint}
\label{sec:orderindep}

Order-independence is the requirement that continuation depends only on \emph{what is known} (the information event), not on \emph{how} it was learned. This is a substantive coherence requirement in sequential/causal entropy settings where conditioning structure and order are central modeling choices \citep{ziebart2010mce,ziebart2013mce}. In the present identified class, the PMI-shaped interaction is pinned by the posterior log-ratio, so requiring a single value function (assumption A2) across update directions forces the two directional representations to agree, yielding an integrability constraint.

\paragraph{Swapped update.}
Rather than assuming the transformation $P(x|y) \mapsto P(x|y,z)$, we swap the direction, i.e. $P(z|y) \mapsto P(z|x,y)$. Define the analogue of \eqref{eq:Jw} for optimizing $\tilde{P}(z\mid y)$ after observing $x$:
\begin{equation}
\label{eq:Ju}
    J_x(\tilde{P}(z\mid y))
    := \sum_z \tilde{P}(z\mid y)\Bigl[
        r_x(z\mid y)
        -\frac{1}{\alpha}\log\frac{\tilde{P}(z\mid y)}{P(z\mid y)}
        + V(z,y,x)
    \Bigr],
\end{equation}
and impose posterior identification for this update as well:
\begin{equation}
\label{eq:post-id-swapped}
    P(z\mid y,x) := \arg\max_{\tilde{P}(z\mid y)} J_x(\tilde{P}(z\mid y)).
\end{equation}
Applying the PMI-shape identity to \eqref{eq:Ju} yields
\begin{equation}
\label{eq:pmi-swapped}
    \log\frac{P(z\mid y,x)}{P(z\mid y)}
    =
    \alpha\Bigl[r_x(z\mid y)+V(z,y,x)-V(y,x)\Bigr]
    = i(z;x\mid y).
\end{equation}
Since A2 imposes event-based coherence, $V(z,y,x)=V(x,y,z)$ and $V(y,x)=V(x,y)$.
Moreover $i(z;x\mid y)=i(x;z\mid y)$ by symmetry of mutual information. Therefore the two update orders
imply the same interaction term (given $y$), yielding the commutativity constraint
\begin{equation}
\label{eq:commute}
    r_z(x\mid y)-V(y,z)
    =
    r_x(z\mid y)-V(x,y).
\end{equation}
Constraint \eqref{eq:commute} reads as an integrability condition:
once interactions are fixed (by PMI under posterior identification),
coherence across update directions forces reward parametrizations across update orders
to be compatible.

\section{Extension to a countable family of variables}

\paragraph{Countable ambient model.}
Let $(X_i)_{i\in\mathbb{N}}$ be a countable collection of discrete random
variables with a joint distribution $P$ on the product space. Let $X,Y,Z$ denote
finite \emph{groups} of coordinates, and write $P(X,Y,Z)$ for the corresponding
marginal. All conditional/marginal notation below is induced by this same
ambient joint $P$.

\paragraph{Minimal well-posedness assumptions.}
Fix a finite triple of coordinate groups $(X,Y,Z)$ and work with the induced
marginal $P(X,Y,Z)$ of the ambient joint on the countable product space. Assume:
\begin{enumerate}[leftmargin=*]
\item \textbf{Positive-mass contexts.}
For every context at which we write $P(\cdot\mid y)$ or $P(\cdot\mid y,z)$ (or the
swapped analogue), assume $P(y)>0$ and $P(y,z)>0$ (resp.\ $P(y,x)>0$) so these
conditionals are defined pointwise.

\item \textbf{Finite log-normalizer (exponential integrability).}
For each $(y,z)$ of interest,
\begin{equation}
\label{eq:countable-Z}
Z(y,z)
:=\sum_x P(x\mid y)\exp\Bigl\{\alpha\bigl[r_z(x\mid y)+V(x,y,z)\bigr]\Bigr\}
<\infty,
\end{equation}
and analogously for the swapped update.
\end{enumerate}

\paragraph{Remark (absolute continuity).}
When all conditionals are induced by the same joint $P$, absolute continuity is
automatic on positive-mass contexts: $P(x\mid y,z)>0\Rightarrow P(x\mid y)>0$ and
$P(z\mid y,x)>0\Rightarrow P(z\mid y)>0$. We do not assume it separately.

\paragraph{Consequence (existence and uniqueness of the soft optimizer).}
Under $Z(y,z)<\infty$, the optimizer exists and is unique, and equals the
exponential tilt
\[
\tilde P^\star(x\mid y;z)
=\frac{P(x\mid y)\exp\{\alpha[r_z(x\mid y)+V(x,y,z)]\}}{Z(y,z)}.
\]
Indeed, for any $q(\cdot\mid y)\ll P(\cdot\mid y)$,
\[
J_z(q)=\frac{1}{\alpha}\log Z(y,z)-\frac{1}{\alpha}\mathrm{KL}\!\left(
q(\cdot\mid y)\,\middle\|\,\tilde P^\star(\cdot\mid y;z)\right),
\]
so $J_z(q)\le \alpha^{-1}\log Z(y,z)$ with equality iff $q=\tilde P^\star$.
All pointwise identities (log-ratio, PMI-shape, gauge invariance, commutativity)
then apply to the marginal triple
$(X,Y,Z)$ exactly as in the finite case.

\section{Discussion}

Our assumptions are deliberately strong, and the payoff is a sharp correspondence between probabilistic conditionals and KL-regularized optimization primitives.

\paragraph{What posterior identification fixes.}
Under posterior identification (A1), the soft optimizer is a genuine conditional under the same joint $P$, so the posterior log-ratio is pinned pointwise.
Equivalently, the identified object is the interaction
$r_z(x\mid y)+V(x,y,z)-V(y,z)$ via \eqref{eq:pmi-shape}. This is a representation
result: PMI is not assumed, it is forced by identifying the optimizer with a
Bayes posterior under one ambient joint.

\paragraph{Why rewards are ambiguous without calibration.}
Because only the interaction is pinned, the decomposition into ``reward'' and
``soft value'' inherits a context-only baseline invariance \eqref{eq:gauge}.
This is the precise obstruction to reading $r_z(x\mid y)$ as a pointwise reward
representation of the posterior update.

\paragraph{Order-robustness as a coherence constraint.}
If we require posterior identification across update directions with a single
continuation event-function (A2), coherence across conditioning orders imposes
the commutativity constraint \eqref{eq:commute}. This is an integrability
condition: once the PMI-shaped interactions are fixed, the direction-indexed
reward families must fit together across update orders.

\paragraph{Implications.}

\emph{Psychology:} In the identified class, the $x$-dependent signal that drives behavioral change under new information $z$ is the conditional log-evidence ratio
$i(x;z\mid y)$ (scaled by $1/\alpha$).
Interpreted as a reward proxy, $i(x;z\mid y)$ quantifies how strongly the agent is ``pulled'' toward or away from outcome $x$ when incorporating $z$ beyond the baseline context $y$.
The gauge \eqref{eq:gauge} formalizes why observed conditionals determine the \emph{relative} propensity shifts but not absolute reward baselines without calibration.

\emph{Economics:} Relative to Savage-style subjective expected utility, the identified class is not a generic preference representation: exact posterior identification ties the utility-relevant interaction to posterior log-likelihood ratios. The agent’s ``as-if payoffs'' are therefore informational rather than free primitives independent of beliefs. The KL term implements an information-processing (or rational-inattention) cost, so preferences over stochastic policies are generally not linear in mixtures unless one enlarges the consequence space to include informational effort. Finally, imposing a single event-based continuation value across update directions yields a path-independence constraint \eqref{eq:commute}, a coherence restriction absent from classical SEU and violated precisely when order effects prevent a single underlying representation \citep{sims2003ri,mackowiak2023rireview}.

\emph{Control:} posterior identification provides a precise boundary for ``control as inference'' under one ambient $P$: if a KL-regularized update is \emph{literally} a Bayes conditional, then the interaction must be PMI-shaped, and design/learning is naturally posed at the level of the identified interaction rather than absolute rewards. Under A2, modular update directions are not independent degrees of freedom; commutativity couples them. Importantly, this does not eliminate design freedom in a chosen update direction: one may specify rewards/terminal values for the operational conditioning order (as in sequential decision-making), but then posterior identification plus A2 pins the compatible representations for alternative conditioning orders via \(\eqref{eq:commute}\).

\paragraph{Limitations.}
The analysis hinges on exact posterior identification under a single ambient
joint and on treating continuation values as specified. Relaxing either
assumption leads back to posterior-like (but not Bayes-identical) soft updates
and reintroduces identifiability gaps; the present note isolates the exact,
pointwise constraints in the fully identified case.

\section{Related Work}
\label{sec:related}

\paragraph{KL-regularized control and soft values.}
KL (relative-entropy) regularization yields exponential-tilt optimizers and
log-partition ``soft'' values, with particularly transparent structure in
linearly-solvable control \citep{todorov2006lmdp} and control--estimation duality
\citep{todorov2008duality}. The variational foundations underlying these
identities are classical \citep{csiszar1975idivergence,donskervaradhan1975asymptoticII,hartmann2017freeenergy,wainwrightjordan2008gm}.

\paragraph{Control as inference and posterior-like updates.}
A broad literature interprets (regularized) control as probabilistic inference,
often via auxiliary graphical models in which optimal policies correspond to
posteriors or variational posteriors \citep{kollerfriedman2009pgm,kappen2009gmi_arxiv,levine2018rci}.
Generalized Bayesian updating further clarifies when loss-tilted posteriors arise
from optimization \citep{bissiri2016generalupdate}. Our setting is strictly more
constrained: we require \emph{exact} posterior identification under a \emph{single}
ambient joint $P$ and study the resulting pointwise consequences.

\paragraph{Reward ambiguity and identifiability.}
It is well known that behavior does not generally identify rewards uniquely:
reward transformations can preserve policies and induce equivalence classes
\citep{ng1999shaping}. Modern identifiability analyses in (entropy-regularized)
inverse RL characterize when reward recovery is possible
\citep{ziebart2008maxentirl,kim2021rewardid,cao2021identifiabilityirl}. Our ``gauge''
viewpoint is driven by posterior identification: the posterior pins a
baseline-invariant interaction, and any choice of pointwise rewards requires an
additional baseline convention beyond posterior information.

\paragraph{Order and coherence.}
Entropy/KL regularization in sequential decision settings motivates careful
specification of conditioning structure and information flow
\citep{neu2017unifiedentropyregmdp,geist2019regularizedmdp,ziebart2010mce,ziebart2013mce}.
Here, coherence across update directions is enforced by requiring a single
continuation event-function, yielding the commutativity constraint
\eqref{eq:commute}.

\section{Conclusion}
\label{sec:conclusion}

We imposed posterior identification (the optimizer is a true conditional under a single joint $P$) and order-independence (a single continuation event-function across directions). Posterior identification forces PMI-shape \eqref{eq:pmi-shape}: the update is pinned pointwise by the posterior log-ratio, hence by $i(x;z\mid y)$, and identifies the interaction $r_z(x\mid y)+V(x,y,z)-V(y,z)$. This exposes a baseline gauge \eqref{eq:gauge}: posteriors do not identify rewards uniquely without a baseline convention. Finally, requiring a single continuation event-function across update directions yields the commutativity constraint \eqref{eq:commute}, coupling the direction-indexed reward parametrizations needed for order-robust representations.

\bibliography{main}
\end{document}